\documentclass[letterpaper]{article} 
\usepackage{aaai2026}  
\usepackage{times}  
\usepackage{helvet}  
\usepackage{courier}  
\usepackage[hyphens]{url}  
\usepackage{graphicx} 
\usepackage{multirow}
\usepackage{natbib}  
\usepackage{caption} 
\title{Imagine-in-Space: Exploring the Frontier of Spatial Intelligence and Reasoning Efficiency in Vision–Language Models}
\author{
    Xiaoxing Lian,
    Aidong Yang,
    Jun Zhu,
    Peng Wang,
    Yue Zhang,
}
\affiliations{
    Telco AI Labs, AsiaInfo Technologies Limited\\
    lianxx@asiainfo.com\\
    yangad@asiainfo.com\\

}

\usepackage{bibentry}

\begin{document}

\maketitle

\begin{abstract}
    Large language models (LLMs) and vision–language models (VLMs), such as DeepSeek-R1, OpenAI o3 and Gemini 2.5 Pro, have demonstrated remarkable reasoning capabilities across logical inference, problem-solving, and decision-making. However, spatial reasoning—a fundamental component of human cognition that includes mental rotation, navigation, and spatial relationship comprehension—remains a significant challenge for current advanced VLMs.
    We hypothesize that imagination—the internal simulation of spatial states—is the dominant reasoning mechanism within spatial world model.  To test this hypothesis and systematically probe current VLM spatial‐reasoning mechanisms, we introduce \textbf{SpatiaLite},a fully synthetic benchmark that jointly measures spatial–reasoning accuracy and \emph{reasoning efficiency}. 
    Comprehensive experiments reveal:  
    (1) Advanced VLMs predominantly rely on linguistic representations for reasoning and ‘imagination’, resulting in significant deficiencies on visual-centric tasks that demand perceptual spatial relation and 3D geometry transformation capabilities like mental rotation or projection prediction. (2) Advanced VLMs demonstrate a profound inefficiency in their current spatial reasoning mechanisms. While VLM like \textit{Gemini 2.5 Pro} can effectively leverage \emph{visual} input to boost both accuracy and token efficiency in collaborative spatial reasoning tasks like \textit{Wood Slide}. The vision input triggers a intuitive vision-driven strategy to identify the primary goal and key steps. (3) We proposed a Imagery Driven Framework (IDF) data synthesis and training framework, which can implicitly construct an internal world model which is critical for VLMs in spatial reasoning. 
    Building on \textbf{SpatiaLite}, this work delineates the spatial reasoning limits and patterns of advanced VLMs, identifies key shortcomings, and informs future advances. Our benchmark code is publicly available at \url{https://github.com/jameslian87v5/SpatialLite}.
\end{abstract}


\section{Introduction}

With the rapid development of vision-language models(VLMs) which have shown significant potential in achieving human-level intelligence of visual perception, planning, understanding and decision-making for complex tasks \cite{wang2025multimodal}. Especially advanced VLMs like o3 \cite{openai2025o3}, Gemini 2.5 Pro \cite{comanici2025gemini} and Claude 3.7 thinking \cite{anthropic2025claude} which have demonstrated surprising reasoning capability in complex tasks like advanced mathematics with visual perception and understanding \cite{deng2025boosting, wang2025mv}. While these VLMs have advanced linguistic intelligence, their visual-spatial intelligence remains unexplored, especially in handling simple reality tasks involve basic spatial reasoning capabilities,such as mental rotation,spatial imagination/prediction and spatial navigation\cite{chen2024spatialvlm, chen2025spatial, lin2025mind}. These capabilities are critical for applications in robotics manipulating and autonomous driving, which can be easily learned by 5 years old child or animals yet remain challenging even for the most advanced VLMs.
\begin{figure}[t]
    \centering
    \includegraphics[width=0.95\columnwidth]{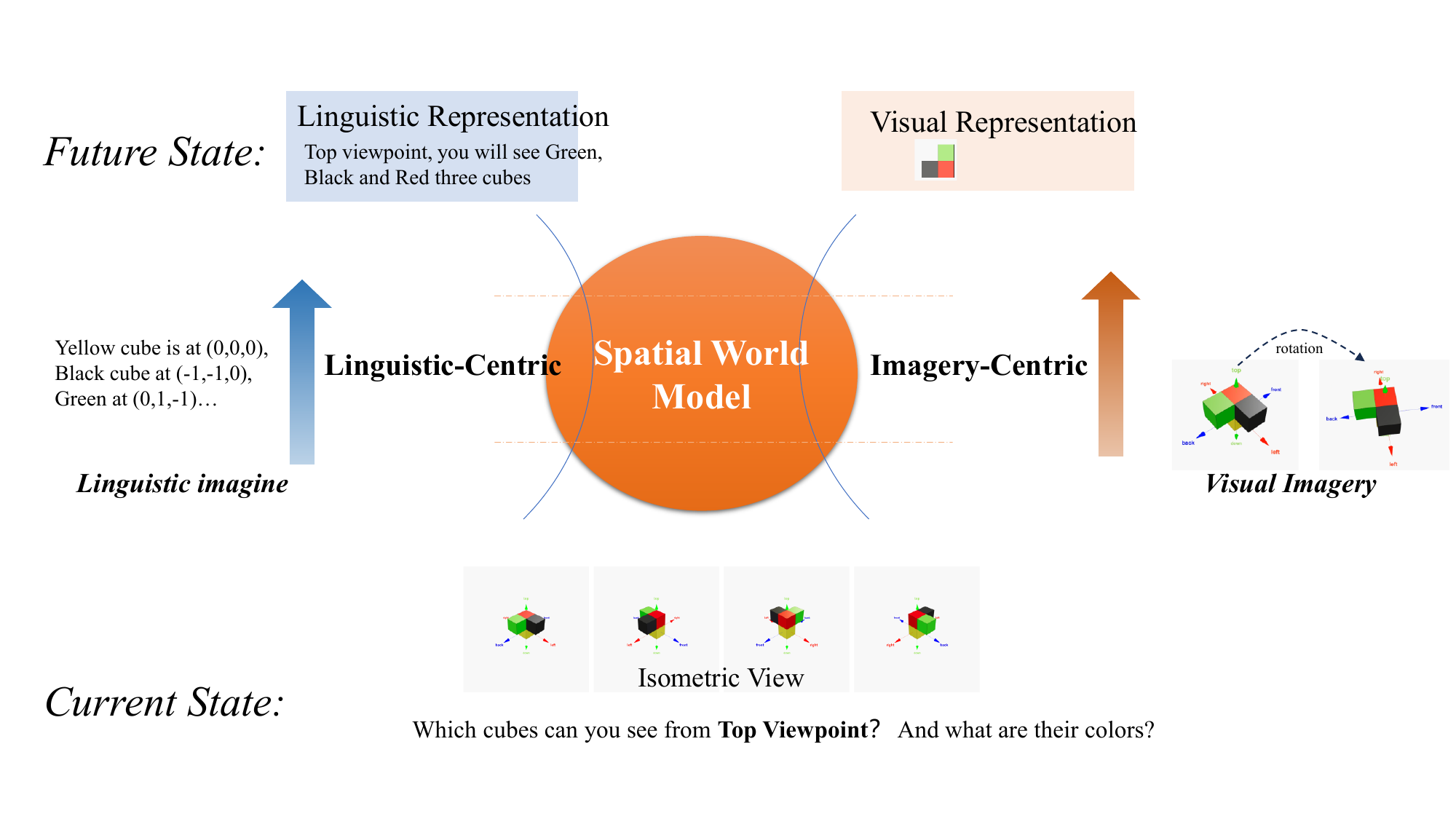}
    \caption{Spatial World Model: Imagination as the core mechanism with two forms - linguistic and visual imagery}
    \label{fig:spatial-world-model}
\end{figure}
To investigate the actual spatial reasoning pattern and capability boundaries of VLMs, we proposed \textbf{SpatiaLite}, a fundamental spatial reasoning benchmark containing simple tasks (Cube rolling, Rubik's cube, Mental rotation, Moving box, Wood slide) representing various dimensions of spatial cognition, from mental rotation to spatial transformation prediction and dynamic path planning.Systematic and comprehensive evaluation of advanced VLMs on \textbf{SpatiaLite} reveals that a large performance gap exists between advanced VLMs and humans in tasks requiring non-linguistical reasoning, and severe inefficiency in tasks requiring dynamic spatial representation (e.g., Moving box, Wood slide), where token usage increases exponentially with transformation complexity. Detailed analysis of VLMs` spatial reasoning behaviour show that the performance inefficiency stems from insufficient spatial transformation and representation capabilities, leading to compensatory over-reliance on linguistic representations. 

To address these limitations, inspired by human visual imagery theory \cite{jeannerod2005visual, xie2020visual} which plays a crucial role in enhancing spatial understanding, spatial working memory processing, and predictive spatial reasoning. Visual imagery aids humans in effectively processing spatial information through simulated visual perception, visual-spatial representation, and enhanced spatial working memory processing. Based on these principles, we proposed  (\textbf{I}magery-\textbf{D}riven \textbf{F}ramework (\textbf{IDF})) data synthesis pipeline and training framework that can generate large scale of high-quality visual imagery data, effectively addressing the data scarcity challenge in construction of spatial world model, which enabling the generation of extensive spatial transformation and imagery data without requiring human annotation. Our main contributions are as follows:

\begin{itemize}
    \item \textbf{Spatial-World-Model Perspective.} Hypothesizing that \emph{imagination} is the dominant mechanism of the spatial world model, we re-examine VLMs from this perspective and introduce \textbf{SpatiaLite}—the first benchmark designed to categorize \emph{visual-centric}, \emph{linguistic-centric}, and \emph{collaborative} spatial reasoning tasks, thus focusing on \emph{spatial working-memory} mechanisms rather than superficial vision–language skills. 
    \item \textbf{Comprehensive VLM Evaluation.} We demonstrate that VLMs primarily leverage \emph{linguistic imagery} for spatial reasoning. This results in a stark performance divide: they achieve superhuman accuracy on \emph{Linguistic-Centric} tasks but fail profoundly on \emph{Visual-Centric} tasks, revealing a core deficit in their visual-spatial representation capabilities.
    \item \textbf{First Efficiency Analysis.}We present the first analysis of spatial reasoning efficiency in VLMs, highlighting a profound inefficiency in their current spatial reasoning mechanisms, where token usage grows exponentially with transformation complexity. Visual input boosts \textit{Gemini 2.5 Pro} on easy tasks by triggering an intuitive, vision-driven strategy that quickly identifies the goal and performs token-efficient heuristic search.
    \item \textbf{Imagery-Driven Framework (IDF).} To construct and enhance the internal spatial world model within VLMs, we propose IDF, a two-stage imagery-distillation pipeline that synthesizes large-scale visual-imagery data. Our experiments validate this approach, demonstrating that IDF significantly improves VLM spatial reasoning.
\end{itemize}

\section{Spatial Intelligence and Categorization}\label{sec:spatial_categorization}
\subsection{Spatial Intelligence and World Model}
With the rapid advancing reasoning capability of LLMs and VLMs, VLMs have begun to exhibit fundamental capabilities in spatial perception, understanding, and transformation. It is essential to explore whether VLMs can serve as spatial world models \cite{yang2025thinking}, and the underlying mechanisms of world models within spatial intelligence through three key aspects \cite{zhang2024spartun3d}: \textbf{Structured representation} – encoding of 3D representations of objects, spatial relationships, object orientations and depth of scenes; \textbf{Transformation prediction} –enable the comprehension and prediction of spatial transformations such as geometric transformations, egocentric-allocentric transformation and perspective projection transformation; \textbf{Strategic planning} – internal simulations of actions and their consequences in spatial environment, enables counterfactual reasoning – exploring "what if" scenarios – which is fundamental for robust planning and decision-making.
\begin{figure*}[t]
    \centering
    \includegraphics[width=0.95\textwidth]{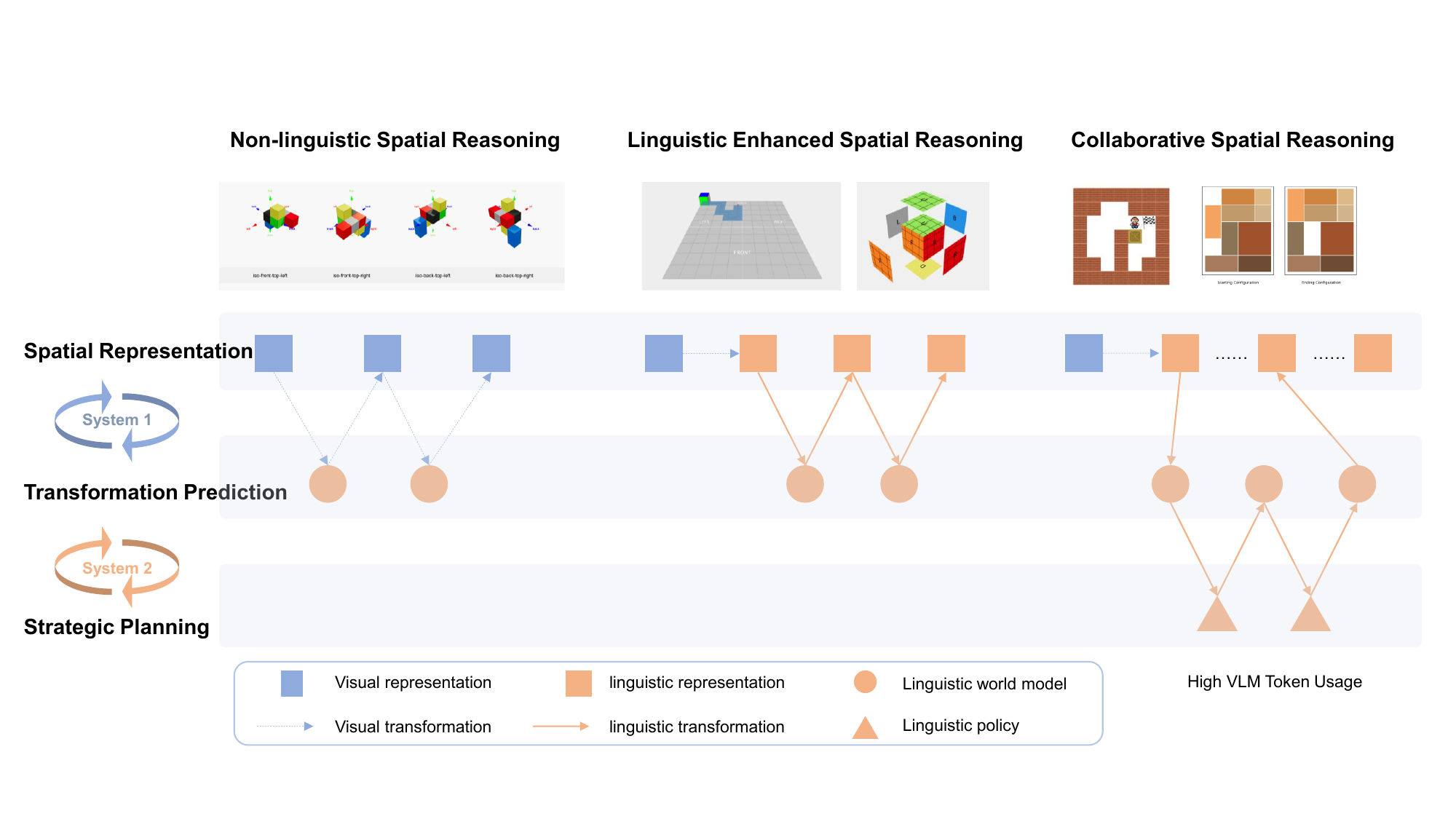}
    \caption{\textbf{SpatiaLite} Benchmark Task Category}
    \label{fig:benchmark_task_category}
\end{figure*}

\textbf{\textit{Inspired by visual imagery theory, we hypothesize that \emph{imagination}—the internal simulation of spatial states—is the dominant reasoning mechanism within spatial world model.}} Current VLMs may realise imagination in two distinct forms:
(i) \textbf{Linguistic imagery} – spatial states and transformations are encoded as symbolic expressions; and VLMs can reason, predict, and plan entirely over this language-level representation.
(ii) \textbf{Visual imagery} – the model relies on visual representations, performing perception, prediction, and decision-making with minimal linguistic mediation, akin to human mental imagery.

\subsection{Spatial Reasoning Task Categorization}
To probe the mechanisms of \textit{spatial world models} \cite{nie2025wmnav} in spatial reasoning, we categorize spatial reasoning tasks into three categories: non-linguistic spatial reasoning, linguistically-enhanced spatial reasoning, and collaborative spatial reasoning.
\begin{itemize}
    \item \textbf{Visual Centric Spatial Reasoning Tasks:} Difficult to linguistically represent or describe such as the geometric transformation of irregular objects, egocentric-allocentric transformations, and perspective projection transformations. We focus on the \textit{Mental Rotation} \cite{shepard1988mental} as a core example of visual-centric spatial reasoning probing.
    \item \textbf{Linguistic Centric Spatial Reasoning Tasks:} Which spatial representations or transformations can be abstracted and described linguistically, allowing for language-guided spatial reasoning. We focus primarily on two representative tasks: \textit{Cube Rolling:} Given an initial cube with colored faces and a sequence of rolls actions, predicting the color of a specified face after all movements. \textit{Rubik's Cube:} Given a 3x3 cube and a sequence of rotation commands, inferring the final state of the cube.
    \item \textbf{Collaborative Spatial Reasoning Tasks:} Which demand the integration of spatial representation, state prediction, and multi-step strategic planning. They involve anticipating future states and iterative problem-solving, with a vast potential state space. We use \textit{Moving Box (Sokoban)} and \textit{Wood Slide (Huarongdao)} \cite{ghosal2024language} as our primary benchmarks for this category.
\end{itemize}

\section{The \textbf{SpatiaLite} Benchmark}
\subsection{Overview}
We present \textbf{SpatiaLite}, a fully synthetic benchmark that enables rigorous, multidimensional evaluation of spatial intelligence in VLMs without risk of data leakage. It focuses on examining VLMs' abilities in structured spatial representation, dynamic transformation understanding, and strategic spatial planning. In all tasks, scenes and annotations are produced by an end-to-end simulation-rendering pipeline, eliminating manual labeling and ensuring unlimited scalability. Its difficulty and scale are highly extensible, enabling precise evaluation of the upper bounds of VLMs’ spatial reasoning capabilities over a broad range of task complexities. \textbf{SpatiaLite} covers five types of tasks: \textit{Mental Rotation}, \textit{Cube Rolling}, \textit{Rubik's Cube}, \textit{Moving Box}, and \textit{Wood Slide}. In terms of evaluation paradigm, we adopts an \textbf{LLM-as-a-parser} protocol and, for the first time, complements accuracy with the novel metric of \emph{reasoning efficiency}, which jointly captures task complexity and the token cost of inference.
\subsection{Task Design and Data Creation}
\begin{figure}[t]
    \centering
    \includegraphics[width=0.9\columnwidth]{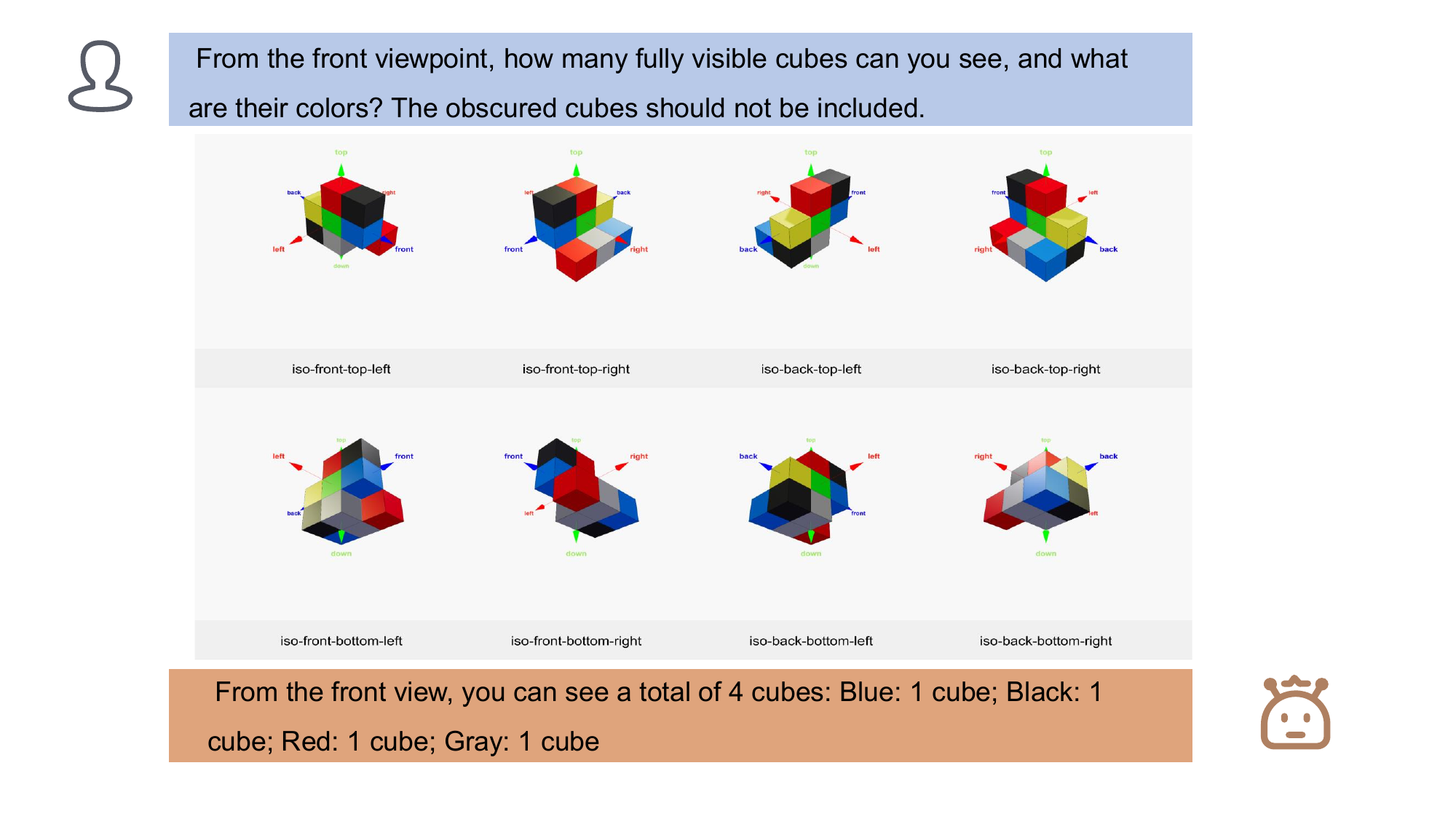}
    \caption{Mental rotation presents VLMs with eight isometric viewpoints of a 3D irregular structure made of colored cubes and ask to infer from one orthogonal viewpoint.}
    \label{fig:mental-rotation-demo}
\end{figure}
\paragraph{Mental Rotation.} Given eight isometric views of an irregular cube assembly (4–13 cubes), the model must recover the full 3-D configuration. Difficulty scales with the number of cubes. Unlike classic single-axis rotation tests, this setting requires multi-view integration and occlusion reasoning.

\paragraph{Cube Rolling.} Given a sequence of rolls, the model must predict the final orientation of a colored cube. Difficulty grows with path tortuosity. The task stresses mental tracking of 3-D rotations beyond single-axis tests. 

\paragraph{Rubik’s Cube.} Starting from a solved 3×3 cube, the model receives a rotation sequence (using U, D, L, R, F, B notation) and must predict the final cube state. Difficulty rises with sequence length, demanding precise mental simulation of coupled face rotations and an expanding state space. Rubik's cube tasks challenge participants to mentally track the evolving configuration of a 3×3 cube through a sequence of rotations. 

\paragraph{Moving Box (Sokoban).} The agent must push boxes to target cells in a grid world on a $6{\times}6$ or $7{\times}7$ grid while avoiding irreversible deadlocks. Difficulty is controlled by the dead-corner density and corridor width, producing levels that range from single-path puzzles to highly branching search spaces. Success requires visuospatial prediction, long-horizon planning, and backtracking under tight space constraints. Levels are generated by a procedural solver-in-the-loop pipeline, ensuring every instance is solvable and annotated with its optimal solution length for benchmarking efficiency.

\paragraph{Wood Slide (Huarong Dao).} Blocks can only slide into adjacent empty spaces and cannot be rotated or lifted. This constraint creates a complex spatial problem: players must carefully sequence moves, often temporarily relocating blocks to create pathways. The task evaluates spatial prediction, analytical assessment, spatial visualization, and problem-solving as players mentally simulate future states to find optimal paths. Limited empty spaces demand precise movement planning and strategic foresight. Same as \textit{Moving box}, we also developed an automated level generation pipeline for solvable puzzles of varying complexity.
\subsection{LLM-as-a-parser Evaluation Design}
To accurately assess VLMs' spatial reasoning, we introduce the \textbf{LLM-as-a-parser} approach. This method tackles random guessing in multiple-choice formats and the non-unique solutions in open-ended tasks. For deterministic tasks ( \textit{Mental rotation, Cube Rolling, Rubik's Cube}), LLMs extract structured data from responses and compare it with ground truth, avoiding chance-level accuracy. For open-ended solution generation tasks ( \textit{Moving box, Wood Slide}), LLMs extract structured action sequences from responses and validate it in the simulation environment. This design thereby enabling objective verification when multiple valid plans exist and allowing the VLM to focus on generating feasible solutions without strict output-format constraints, which improves the generality and effectiveness of evaluation.
\subsection{Difficulty Design}
\begin{figure}[t]
    \centering
    \includegraphics[width=0.9\columnwidth]{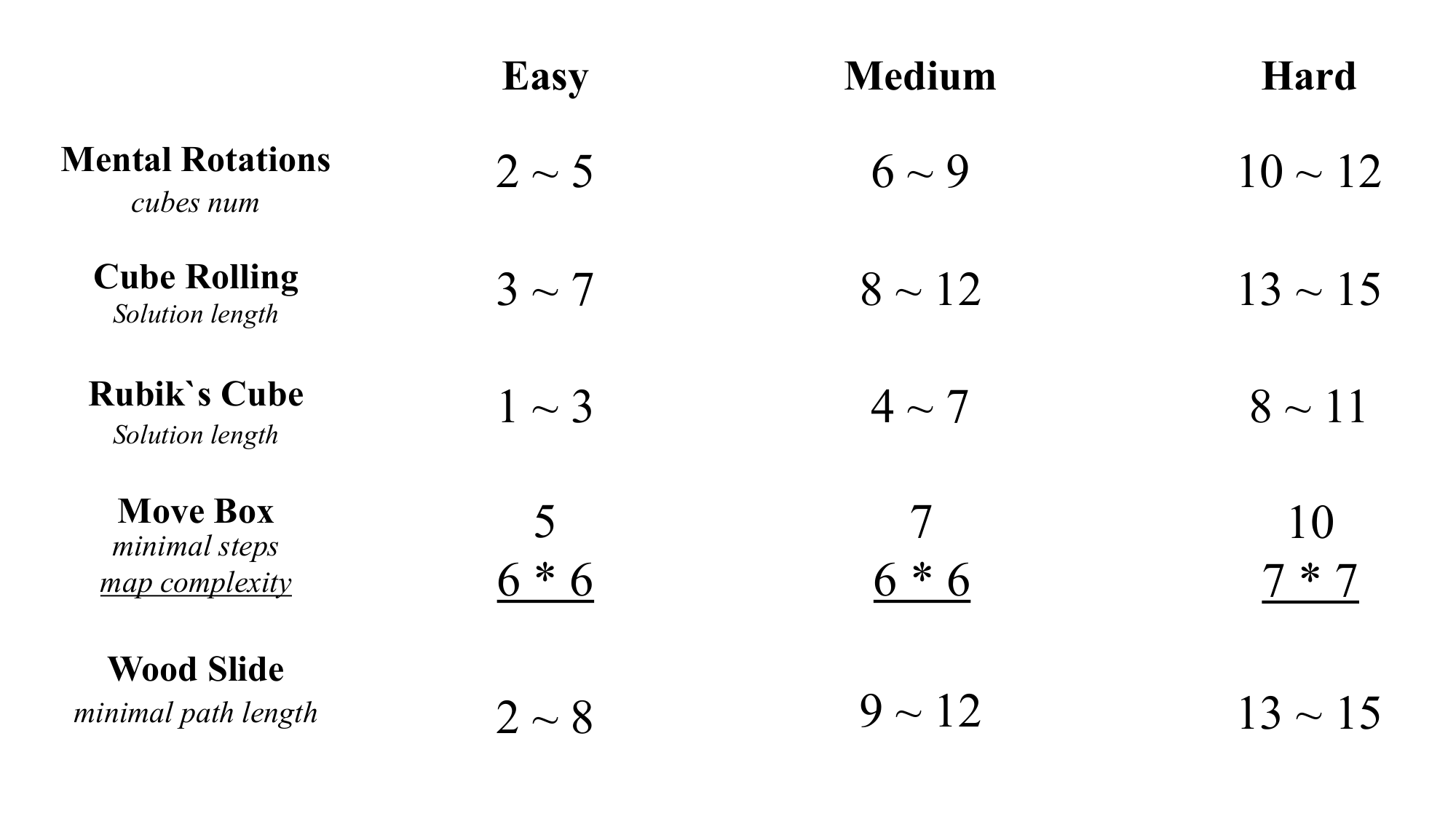}
    \caption{Complexity setting for each task and difficulty level.}
    \label{fig:complexity-setting}
\end{figure}
A comprehensive evaluation of VLMs' spatial reasoning, each employing corresponding difficulty metrics for complexity quantification standards (as illustrated in Figure \ref{fig:complexity-setting}).For \textit{Mental Rotation} tasks, complexity is primarily determined by the number of cubes, increasing the number of cubes proportionally elevates the cognitive demands of viewpoint mapping and occlusion reasoning. For \textit{Cube Rolling}, complexity is determined by the tortuosity of the rolling path. For \textit{Rubik's Cube}, task complexity scales with the length of the rotation sequence, as each additional move expands the state space and intensifies the working-memory demands of mentally tracking cube transformations. For \textit{Moving Box}, complexity is assessed through a multi-factor metric that weights dead-corner count, narrow-passage prevalence, and obstacle density, thus reflecting trap density and manoeuvre constraints rather than relying solely on shortest-path or Manhattan distance. For \textit{Wood Slide}, difficulty is primarily quantified by the optimal solution path length from initial to target state.

\section{Experimental Evaluation}

\subsection{Evaluation Setup}
\paragraph{Benchmark Models.} The comparative experiments accessed proprietary models via their official APIs and employed LMDeploy to evaluate open-source models. This study benchmarks reasoning-enhanced VLMs which have undergone reinforcement learning post-training processes against their non-enhanced counterparts. Many of these models claim state-of-the-art performance on various multimodal benchmarks, notably OpenAI’s GPT-4o, o1, and o4-mini; Google’s Gemini 2.5 Pro and Gemini 2.5 Flash; and the Qwen\_2.5\_VL family \cite{bai2025qwen2} with Qwen\_2.5\_VL-32B and Qwen\_2.5\_VL-7B. We also include the InternVL\_3 series \cite{zhu2025internvl3} InternVL\_3\_9B and InternVL\_3\_38B. Finally, we adopt DeepSeek-R1 \cite{guo2025deepseek} as a baseline for linguistic centric spatial reasoning capabilities on tasks which can be symbolized or text-described.
\paragraph{Human Level Performance.} We randomly sampled 30 questions per subtask in proportion to each difficulty tier (150 questions in total) and tested ten graduate-level volunteers independently within a two-minute time limit, \emph{using scratch paper to augment their working memory}. For VLMs, we imposed an analogous constraint by limiting each model’s maximum output tokens. 
\begin{table*}[ht]
    \caption{Accuracy (\%) on \textbf{SpatiaLite} benchmark by task.
             Models are grouped into Baselines, Reasoning-enhanced VLMs, and Non-enhanced VLMs.}
    \label{tbl:task-accuracy}
    \centering
    \setlength{\tabcolsep}{6pt}
    \begin{tabular}{lccccc}
      \hline
      \textbf{Model} & \textbf{Mental Rot.} & \textbf{Cube Rolling} &
      \textbf{Rubik's Cube} & \textbf{Move Box} & \textbf{Wood Slide} \\
      \hline
      \multicolumn{6}{c}{\textit{Baselines}}\\
      \hline
      Chance Level (Random)   & -- & 16.7 & 16.7 & -- & -- \\
      Chance Level (Frequency)   & -- & 18.5 & 26.6 & -- & -- \\
      Human Level            & 100.0 & 85.3 & 92.3 & 99.2 & 98.6 \\
      DeepSeek-R1  & -- & 75.6 & 68.4 & 77.3 & 71.5 \\
      \hline
      \multicolumn{6}{c}{\textit{Reasoning-Enhanced VLMs}}\\
      \hline
      o1$^*$   & 7.5 & 97.5 & 77.6 & 91.4 & 66.8 \\
      o4-mini$^*$   & 6.4 & 98.3 & 74.5 & 89.7 & 58.4 \\
      Gemini 2.5 Pro         & 20.5 & 75.0 & 67.3 & 30.6 & 52.3 \\
      Gemini 2.5 Flash       & 4.5 & 62.3 & 41.5 & 22.6 & 37.4 \\
      \hline
      \multicolumn{6}{c}{\textit{Non-enhanced VLMs}}\\
      \hline
      GPT-4o                 & 4.7 & 51.3 & 53.8 & 3.6 & 4.9 \\
      Gemini 2.0 Flash       & 3.8 & 17.5 & 40.2 & 1.5 & 0 \\
      Qwen\_2.5\_VL\_72B    & 5.6 & 32.5 & 31.1 & 0 & 0 \\
      Qwen\_2.5\_VL\_32B   & 2.2 & 34.1 & 32.4 & 0 & 0 \\
      Qwen\_2.5\_VL\_7B    & 4.3 & 12.5 & 20.1 & 0 & 0 \\
      InternVL\_3\_78B      & 6.8 & 38.6 & 33.4 & 1.5 & 2.8 \\
      InternVL\_3\_38B      & 3.5 & 32.6 & 28.3 & 0 & 2.5 \\
      InternVL\_3\_9B       & 3.4 & 17.4 & 31.3 & 0 & 0 \\
      \hline
    \end{tabular}
    
    $^*$medium effort
\end{table*}
\subsection{Overall Performance}
Table~\ref{tbl:task-accuracy} reports the performance of various VLMs on the \textbf{SpatiaLite} benchmark, categorizing the models into three groups: Baselines, Reasoning-enhanced VLMs, and Non-enhanced VLMs.

\paragraph{Visual-Centric Spatial Reasoning Remains Challenging.}
Current advanced VLMs perform poorly on \textit{Mental rotation} tasks: with \textit{Gemini 2.5 Pro} leading top performance at 20.5\%, whereas all other models scores below 10\%. By contrast, human participants achieve nearly 100 \% accuracy on the same task. This gap highlights a key weakness in current VLMs: they still lack the visuospatial imagination which is essential to mentally transform or predict  spatial representation from alternate viewpoints. This transformation can not be processed solely by linguistical, especially for the complex spatial structures. Analysis of \textit{Gemini 2.5 Pro}’s reasoning trajectory shows that it first maps the spatial structure to an explicit spatial coordinate system and then performs purely linguistic reasoning over these spatial coordinates. These process succeeds for very simple layouts but collapses for moderately complex layouts.

\paragraph{Reasoning-Enhanced VLMs Significantly Outperform  Non-Enhanced Counterparts.}
Reasoning-enhanced VLMs markedly outperform their non-enhanced counterparts on linguistically centric spatial-reasoning tasks. In \textit{Cube Rolling}, \textit{o4-mini} and \textit{o1} achieve 98.3 \% and 97.5 \% accuracy, respectively, surpassing human performance and demonstrating strong long-context reasoning ability. On the more complex \textit{Rubik’s Cube} task, they approach human-level accuracy, while on \textit{Move Box} and \textit{Wood Slide} the gap widens further. Non-enhanced models rarely solve these puzzles and thus score near zero; their failures mainly arise from spatial-transformation or prediction errors, such as mismapping faces in \textit{Cube Rolling} and \textit{Rubik’s Cube}, violating basic movement constraints in \textit{Move Box} and \textit{Wood Slide}, or reasoning chain collapses in the long-horizon sequence-tracking. Whereas reasoning-enhanced VLMs frequently succeed and achieve high scores. 

\paragraph{Open-Source VLMs Remain at Chance-Level Accuracy.}
A clear performance gap exists between proprietary and open-source VLMs: the latter hover at chance level on \textit{Rubik’s Cube} and fail almost entirely on \textit{Move Box} and \textit{Wood Slide}, underscoring their limitations on challenging spatial-reasoning tasks.
\begin{figure}[t]
    \centering
    \includegraphics[width=0.8\linewidth]{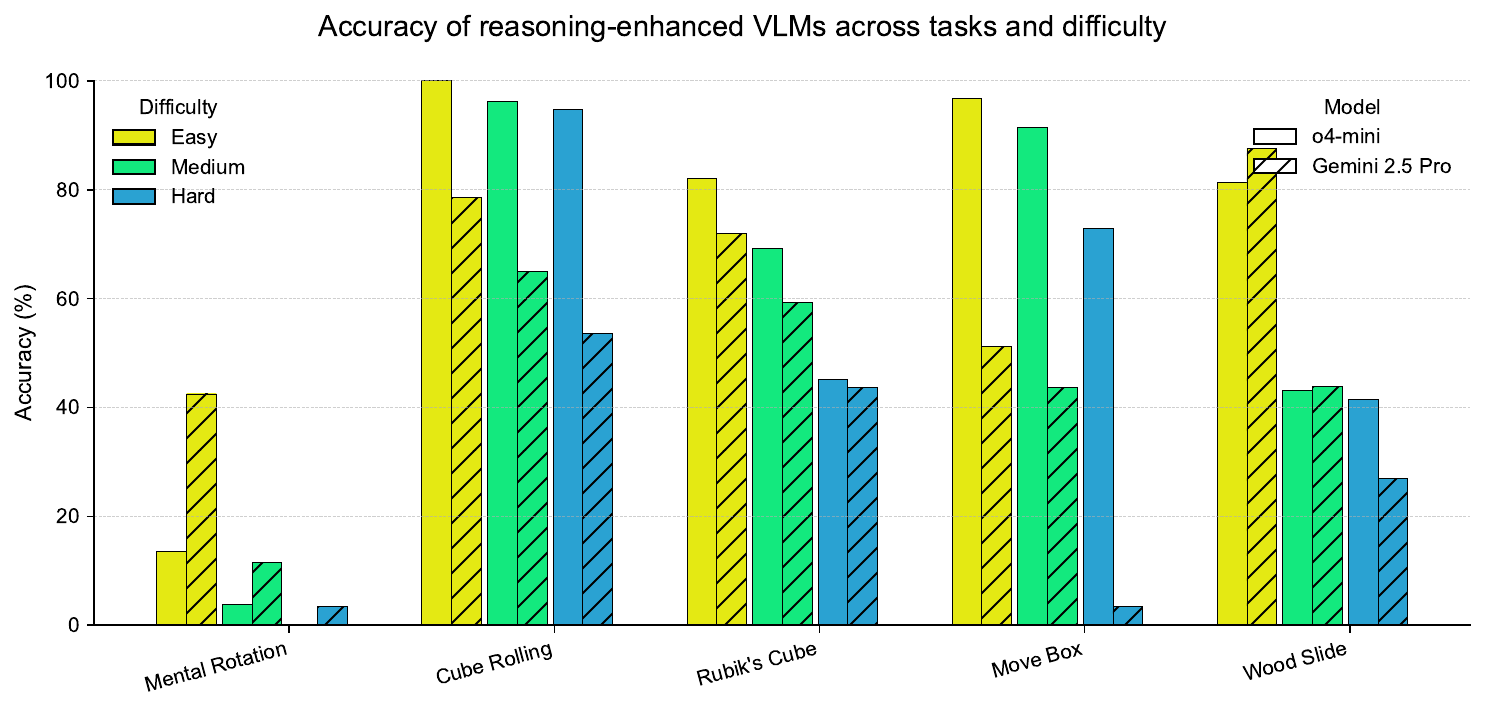}
    \caption{Accuracy comparison across different tasks and difficulty levels}
    \label{fig:accuracy_comparison}
\end{figure}

\begin{figure}[t]
    \centering
    \includegraphics[width=0.8\linewidth]{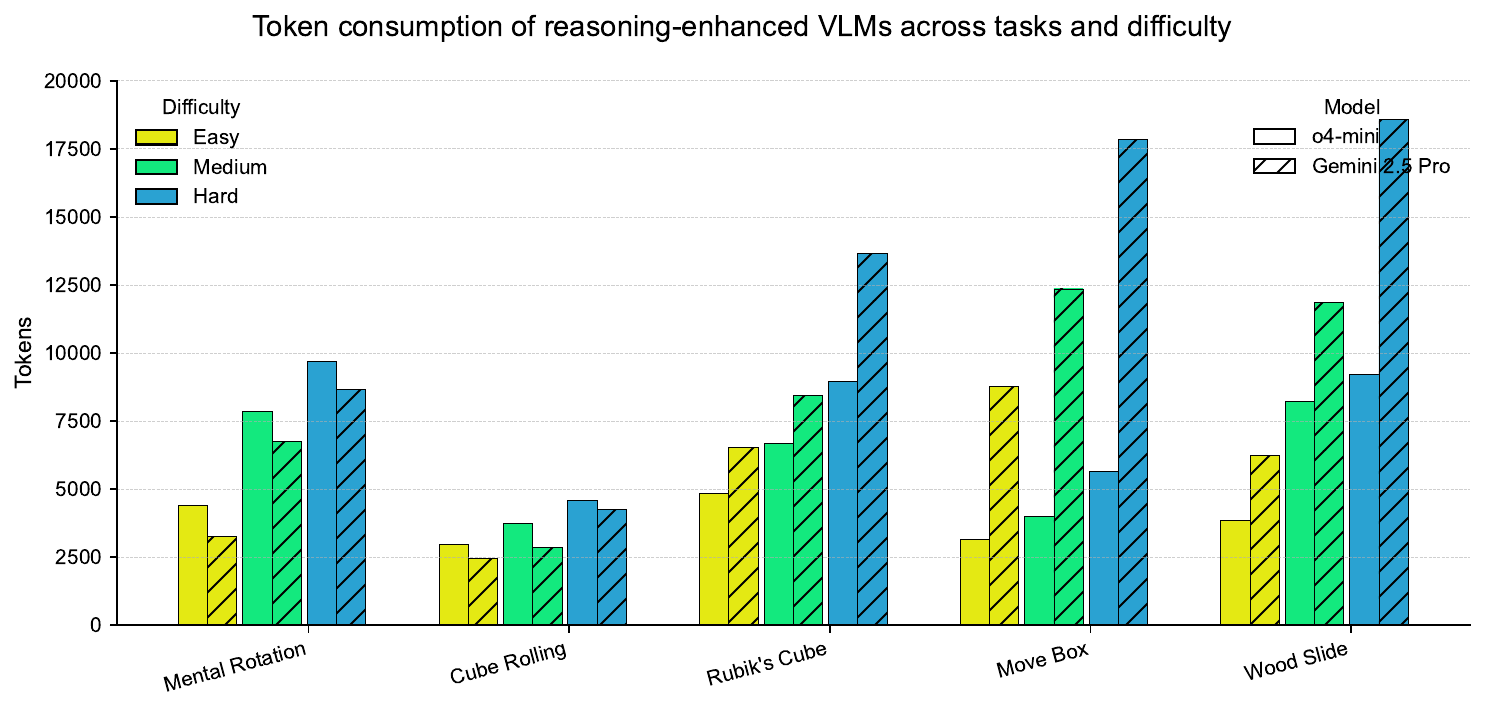}
    \caption{Token consumption comparison across different tasks and difficulty levels}
    \label{fig:token_usage}
\end{figure}



\subsection{Capability Boundaries and Efficiency Analysis}
Both \textit{o4-mini} and \textit{Gemini 2.5 Pro} exhibit trend of declining accuracy and increasing token usage as difficulty rises. In \textbf{Mental Rotation}, \textit{Gemini 2.5 Pro} attains 43\% accuracy on simple layouts containing only 2–5 cubes, but droping sharply as more cubes added. \textit{o4-mini} is virtually unable to solve this task at any difficulty, underscoring the fragility of current advanced VLMs on vision-centric spatial reasoning. In contrast, \textit{o4-mini} attains 94.5\% accuracy on the \textbf{Cube Rolling} \textit{Hard} setting, indicating that the benchmark no longer challenges the model on this task. On \textbf{Move Box}, \textit{o4-mini} still reaches 73.5\% at \textit{Hard} difficulty, while \textit{Gemini 2.5 Pro} manages only marginal progress and is almost unable to solve tasks at this level. This two models perform comparably in \textbf{Wood Slide}. From efficiency perspective, both VLMs generate far more tokens on these spatial tasks than on typically mathematical-reasoning tasks. They both demonstrate ultra inefficiency (average more than 10,000 tokens) on the  \textit{Hard} setting, exceeding on complex mathematical-reasoning problems. Overview, \textit{o4-mini} shows superior robustness and efficiency on linguistic-centric and collaborative spatial-reasoning tasks, whereas both models remain weak on visual-centric problems.  In \textbf{Mental Rotation}, failures are dominated by perception-level deficiencies: the models struggle to maintain a consistent internal spatial representation of object under viewpoint changes. In \textbf{Rubik's Cube}, errors arise chiefly from transformation-level. By contrast, in \textbf{Move Box} and \textbf{Wood Slide} the prevailing errors are strategic level. On the \textit{Hard} tasks that call for deliberate strategic manoeuvres, the VLMs lacking tactical insight and failing to formulate strategic planning, trapping in unproductive moves and causing token usage up sharply.

\section{Modality Analysis: How Visual Input Shape Efficiency and Accuracy}
To identify the main contribution of visual and linguistic signals in spatial reasoning, we follow prior work \cite{wang2024picture} and construct three input modalities terms: VQA, which provides the image plus a terse textual prompt; TQA, which offers only a symbolic textual description; and VTQA, which pairs the image with a detailed symbolic description. Note that within the \textsc{SpatiaLite} benchmark we adopt the VTQA
setting for all tasks \emph{except} \textit{Mental Rotation}, which is kept
in VQA form because providing the cube coordinates would nearly reveal the
answer.
\begin{table*}[ht]
    \caption{The effect of different input modalities on the performance and efficiency of VLMs}
    \label{tbl:modality-comparison}
    \centering
    \setlength{\tabcolsep}{6pt}
    \begin{tabular}{llcccccc}
      \hline
      & & \multicolumn{2}{c}{\textbf{TQA}} & \multicolumn{2}{c}{\textbf{VQA}} & \multicolumn{2}{c}{\textbf{VTQA}} \\
      \cline{3-4} \cline{5-6} \cline{7-8}
      \textbf{Task} & \textbf{Model} & \textbf{Acc.$\uparrow$} & \textbf{Token$\downarrow$(k)} & \textbf{Acc.$\uparrow$} & \textbf{Token$\downarrow$(k)} & \textbf{Acc.$\uparrow$} & \textbf{Token$\downarrow$(k)} \\
      \hline
      \multirow{2}{*}{Wood Slide} 
      & o4-mini        & \textbf{61.2} & 7.1 & 4.5 & 7.6 & 58.4 & \textbf{6.7} \\
      & Gemini 2.5 Pro & 48.5 & 12.3 & 50.6 & \textbf{9.5} & \textbf{52.3} & 11.9 \\
      \hline
      \multirow{2}{*}{Rubik's Cube} 
      & o4-mini        & 74.3 & 6.9 & 66.3 & 6.8 & \textbf{74.5} & \textbf{6.6} \\
      & Gemini 2.5 Pro & 66.5 & 9.1 & 63.2 & \textbf{8.7} & \textbf{67.3} & 8.9\\
      \hline
    \end{tabular}
\end{table*}
\paragraph{No benefit from vision input for \textit{o4-mini}.}
Consistent with prior findings \cite{wang2024picture}, o4-mini performs best under the text-only TQA setting. Adding visual input (VTQA) brings no benefit, and the vision-only VQA condition even harms performance—on Wood Slide its accuracy drops to almost zero. 
These results indicate that o4-mini’s spatial reasoning is dominated by linguistic processing, with negligible positive contribution from visual features.
\begin{figure}[t]
    \centering
    \includegraphics[width=0.75\linewidth]{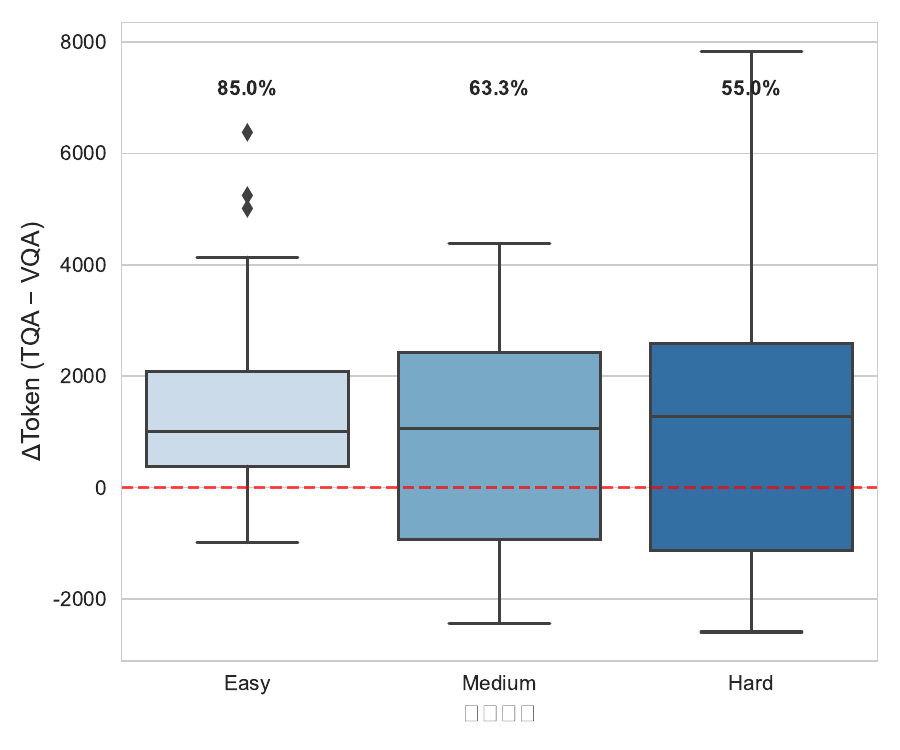}
    \caption{$\Delta$Token (TQA -- VQA) box plot on \textit{Wood Slide} across difficulty levels for \textit{Gemini 2.5 Pro}}
    \label{fig:vqa_tqa_token_comparison}
\end{figure}
\paragraph{Vision input boosts accuracy and efficiency for \textit{Gemini 2.5 Pro}}
Visual input provides significant benefits for Gemini 2.5 Pro. The VTQA testing yields the highest accuracy, whereas the VQA setting achieves better accuracy performance with that of TQA while showing markedly higher efficiency. This pattern is more obviously in the Wood Slide task, suggesting \textit{Gemini 2.5 Pro} can effectively leverage visual representations for spatial perception, understanding and transformation in its reasoning process. 
\paragraph{Why VQA Is More Efficient for \textit{Gemini 2.5 Pro}.}
To explore the underlying reasons, we analyze the trajectories of Gemini 2.5 Pro in VQA and VTQA settings across different difficulty levels. With vision and text impacting the process in complementary ways. We randomly sampled 100 easy, 50 medium, and 50 hard samples on which both VQA and TQA correct solutions. At the easy difficulty, even the lower quartile of $\Delta$Token (TQA -- VQA) remains above zero, confirming a pronounced efficiency advantage for VQA. As task complexity increases, this visual advantage narrows and the variance widens. In complex spatial reasoning process, VLMs need lingusitic signals to organize inference steps to simulate and actively predict action consequences.
VQA formulates heuristic, vision-driven strategies: the VLM leverages initial vision-driven observation to identify the primary goal and corresponding actions first, then executes a trial-and-error and heuristic search which is particularly effective for easy puzzles with fewer tokens. In contrast, TQA relies on symbolic descriptions to identify critical blocks and determine the goal, then performs a systematic, abstract sub-goal decomposition. We posit that this heuristic and vision-driven reasoning enables VQA to achieve goal more quickly on easy tasks; as complexity and the search space expands, leading to VQA’s efficiency advantage narrows and variance widens.

\section{Constructing Internal World Models through Imagination}
We see that the superior spatial reasoning in proprietary VLMs likely stems from their ability to maintain a dynamic internal representation of space or so called internal world model. This representation is linguistic reasoning but grounded in visual signals, enabling predictive simulations crucial for effective planning and forecasting. 
By comparing mental prediction against real environment feedback, human iteratively learn the governing mechanisms of spatial transformations and relationships, which effectively constitutes our internal world model. 
Hence, we propose a two-stage SFT pipeline. The first stage, \textbf{Imagery-Distillation Stage} is desgined to build an implicit internal spatial world model for VLMs, which employ a random-walk simulation process to generate data at scale for this stage. For example, we prompt to predict the cube faces after an ‘R U’ rotation in Rubik’s Cube or update the puzzle layouts after given action sequence in Move Box. Each training sample includes the corresponding thinking process for the prediction which is distilled from \textit{DeepSeek-R1 and Gemini 2.5 Pro}. Finally, we obtain 20 k samples across five tasks. \textbf{Reasoning-Distillation Stage:} Next, the VLM is further fine-tuned on a collected 5k correct trajectories which are obtained from VLM`s successful reasoning of \textit{o4-mini and Gemini2.5 Pro} and pseudo-reasoning paths reconstructed from ground-truth solutions. The training data is carefully filtered to avoid overlap with the benchmark dataset.
\begin{table}[!htbp]
    \caption{Accuracy (\%) of Qwen-2.5-VL-7B under different SFT settings.  
             "Original" denotes the off-the-shelf checkpoint,  
             "RD" uses Reasoning Distillation only,  
             "ID\,+\,RD" employs the proposed two-stage Imagery+Reasoning Distillation.}
    \label{tbl:id-rd-comparison}
    \centering
    \small
    \setlength{\tabcolsep}{3pt}
    \begin{tabular}{lccccc}
      \hline
      \textbf{Model Setting} & \textbf{Mental} & \textbf{Cube} & \textbf{Rubik's} & \textbf{Move} & \textbf{Wood} \\
      \textbf{} & \textbf{Rot.} & \textbf{Rolling} & \textbf{Cube} & \textbf{Box} & \textbf{Slide} \\
      \hline
      Original (no SFT) & 4.3 & 12.5 & 20.1 & 0 & 0 \\ 
      RD (reasoning-only) & 5.6 & 26.4 & 32.4 & 0 & 0 \\
      ID + RD (two-stage) & \textbf{7.5} & \textbf{42.3} & \textbf{44.7} & 0 & 0 \\
      \hline
    \end{tabular}
    \footnotesize{\textit{Note:} "Original" = off-the-shelf model, "RD" = Reasoning Distillation only, "ID+RD" = two-stage Imagery+Reasoning Distillation.}
\end{table}
\paragraph{Imagery distillation helps building internal world model.}
We conducted the experiments using Qwen-2.5-VL-7B through Xtuner using 8 NVIDIA A800 GPUs which freezen the vision encoder and MLP-based Vision-Language Merger of Qwen 2.5 VL Model Architecture. Table XXX compares two settings: RD (reasoning-distillation only) and the two-stage ID + RD SFT. The results demonstrate the ID + RD two stage SFT achieves higher accuracy than chancel level on cube rolling and rubiks cube tasks. And two stage SFT outperform RD alone notably on Rubik’s Cube, where the model learns the cube face transformation mechanics and can identify unaffected faces with high precision. On tasks Move Box and Wood Slide, the two-stage SFT improve the accuracy of board layout prediction. The results suggesting the imagery stage helps for implicitly constructing an internal world model, particularly in understanding and predicting spatial transformations, such as the face mapping mechanics in Cube Rolling and Rubik's Cube. But still struggles in spatial perception and maintaining a stable and persistent reasoning trajectory in move box and wood slide. In the future, we will explore the integration of RL-based Training to further improve the model's performance.

\section{Related Work}
Apart from the analysis of spatial intelligence in Section~\ref{sec:spatial_categorization}, we futher review related work in the following aspects:
\paragraph{Spatial Reasoning of VLMs.}
A predominant paradigm in current VLMs is to construct reasoning chains primarily within the language space \cite{zhang2025spartun3dsituatedspatialunderstanding, cheng2024spatialrgpt, chen2025integratingchainofthoughtmultimodalalignment}, relying solely on the language modality for subsequent information integration and logical deduction. Some work focus on native spatial reasoning training pipeline from directly integration of 3D representations, such as 3 RGB-D Images \cite{ma2024spatialpin, 10658310, 10657856}, depth maps, and point clouds. \cite{deng20253dllavageneralist3dlmms, fei2024kestrel, liu2024pointmllm}  These models exhibit significant 3D understanding but struggle in spatial reasoning as they lack the ability for active perception or mental simulation of spatial dynamics. Other work focus on leveraging visual representation to active perception and mental simulation of spatial state into the reasoning process. \cite{li2025imaginereasoningspacemultimodal, shao2024visual}
\paragraph{Constructing Internal World Models.}
There are multiple ways training internal world models in VLMs. Some employ Supervised Fine-Tuning (SFT) with chain-of-thought data \cite{yao2024mulberryempoweringmllmo1like, shao2024visualcotadvancingmultimodal}. Many approaches further refine models through a two-stage process that combines SFT with Reinforcement Learning (RL) techniques, to better align with desired outcomes \cite{wang2025enhancingreasoningabilitymultimodal,liu2025visualrftvisualreinforcementfinetuning, deng2025boostinggeneralizationreasoningvision}. 

\section{Conclusion and Future Work}
We introduced \textbf{SpatiaLite}, a benchmark designed to evaluate both accuracy and efficiency of spatial reasoning. Our experiments reveal that current VLMs are profoundly inefficient and largely fail at visual-centric tasks due to an over-reliance on linguistic processing. To address this, we proposed the Imagery-Driven Framework (IDF), a two-stage training pipeline that significantly improves spatial reasoning by implicitly constructing a internal world model. Looking ahead, we will expand our benchmark with more diverse tasks and explore the integration of RL-based training, data preparation, and model architecture to further improve the model's performance.
\bibliography{imagine_in_space}


\end{document}